\documentclass[runningheads]{llncs}
\usepackage[T1]{fontenc}
\usepackage{multirow}
\usepackage{amsmath}
\usepackage{bbm}
\usepackage{amssymb}
\usepackage{graphicx}
\usepackage[table]{xcolor}
\usepackage{hyperref,url}
\usepackage{color}

\begin{document}
\title{Formulations and scalability of neural network surrogates in nonlinear
optimization problems}
\titlerunning{Formulations and scalability of neural network surrogates}
\author{
  Robert B. Parker\inst{1}      %\orcidID{0000-1111-2222-3333}
  \and Oscar Dowson\inst{2}     %\orcidID{1111-2222-3333-4444}
  \and Nicole LoGiudice\inst{3} %\orcidID{2222-3333-4444-5555}
  \and Manuel Garcia\inst{1}    %\orcidID{2222-3333-4444-5555}
  \and Russell Bent\inst{1}     %\orcidID{2222-3333-4444-5555}
}
\authorrunning{R. Parker et al.}
\institute{
  Los Alamos National Laboratory, Los Alamos, NM 87544, USA
  \email{\{rbparker,mjgarcia,rbent\}@lanl.gov}\\
  \and Dowson Farms\\
  \email{oscar@dowsonfarms.co.nz}\\
  \and Texas A\&M University, College Station, TX 77843, USA\\
  \email{nicolelogiudice30@tamu.edu}
}
\maketitle              % typeset the header of the contribution
\begin{abstract}
  We compare full-space, reduced-space, and gray-box formulations for representing
  trained neural networks in nonlinear optimization problems.
  We test these formulations on a transient stability-constrained,
  security-constrained alternating current optimal power flow (SCOPF) problem where the
  transient stability criteria are represented by a trained neural network surrogate.
  Optimization problems are implemented in JuMP and trained neural networks are
  embedded using a new Julia package: MathOptAI.jl.
  To study the bottlenecks of the three formulations, we use neural networks
  with up to 590 million trained parameters.
  The full-space formulation is bottlenecked by the linear
  solver used by the optimization algorithm, while the reduced-space formulation
  is bottlenecked by the algebraic modeling environment and derivative computations.
  The gray-box formulation is the most scalable and is capable of solving with the
  largest neural networks tested.
  It is bottlenecked by evaluation of the neural network's outputs and their derivatives,
  which may be accelerated with a graphics processing unit (GPU). Leveraging the gray-box
  formulation and GPU acceleration, we solve our test problem with our largest neural
  network surrogate in 2.5$\times$ the time required for a simpler SCOPF problem without
  the stability constraint.
\keywords{Surrogate modeling  \and Neural networks \and Nonlinear optimization}
\end{abstract}

\section{Introduction}

Nonlinear local optimization is a powerful tool for engineers and
operations researchers for its ability to handle accurate physical models,
respect explicit constraints, and solve large-scale problems \cite{biegler2010}.
However, it is often the case that practitioners wish to include components that do
not easily fit into the differentiable and algebraic frameworks of nonlinear optimization.
Examples include when a mechanistic model
is not available \cite{thebelt2022},
is time-consuming to simulate \cite{omlt},
or renders the optimization problem too complicated to solve reliably \cite{bugosen2024}.

A recent trend has been to replace troublesome components by a trained neural network
surrogate and then embed the trained neural network into a nonlinear optimization model \cite{lopezflores2024}.
Several open-source software packages, e.g., OMLT \cite{omlt} and gurobi-machinelearning
\cite{gurobimachinelearning}, make it easy to embed neural network models in Python-based
modeling environments for nonlinear optimization.
A trained neural network may be represented in an optimization problem with different formulations,
e.g., \textit{full-space} and \textit{reduced-space} formulations
\cite{omlt,schweidtmann2019deterministic}, which have been compared by Kilwein \cite{kilwein2023}
for a security-constrained AC optimal power flow (SCOPF) problem.
%We introduce these more fully in Section~\ref{sec:algebraic}, but briefly: the \textit{full-space}
%formulation adds new variables and constraints for every layer in the neural network,
%whereas the \textit{reduced-space} formulation adds a single constraint that represents
%all the layers. We point to the recent work of \cite{kilwein2023}, which uses OMLT
%to compare these two approaches in the context of a similar security-constrained AC-OPF
%problem.
A third approach, first suggested by Casas \cite{casas2024}, is called
a \textit{gray-box} formulation, in which function and derivative evaluations of the
surrogate are handled by the neural network modeling software (here, PyTorch \cite{paszke2019pytorch}),
rather than the algebraic modeling environment (here, JuMP \cite{jump1}).

While full-space, reduced-space, and gray-box formulations have been compared in \cite{casas2024},
the bottlenecks of these formulations have not been carefully identified.
This paper profiles these three formulations
of a neural network predicting transient feasibility
in an SCOPF problem.
We demonstrate that the gray-box formulation is the most scalable, and
that it can naturally take advantage of GPU acceleration
built into PyTorch. We find that the full-space formulation is bottlenecked
by the solution of a linear system of equations, and the reduced-space formulation
is bottlenecked by JuMP and its automatic differentiation system.
Our work provides a clear benchmark and direction for future work.
%Our work differs to that of \cite{casas2024} in that we study a different application
%domain and we provide more detailed timing information to understand why each
%formulation has a different performance characteristic.
Additionally, we provide MathOptAI.jl, a new open-source library for
embedding trained machine learning predictors into optimization models built
with JuMP \cite{jump1}. MathOptAI.jl is available at \url{https://github.com/lanl-ansi/MathOptAI.jl}
under a BSD-3 license. 

\section{Background}

\subsection{Nonlinear optimization}\label{sec:nonlinear_opt}
We study nonlinear optimization problems in the form given by
Equation \eqref{eqn:nlopt}:
\begin{equation}
  \min_x f(x) \text{ s.t. } g(x)=0; ~x \geq 0.
  %\begin{array}{cl}
  %  \displaystyle \min_{x} & f(x) \\
  %  \text{s.t.} & g(x) = 0\\
  %  & x \geq 0 \\
  %\end{array}
  \label{eqn:nlopt}
\end{equation}
%We consider lower bounds of zero for simplicity, but general upper and lower
%bounds can be handled easily.
%Problems with general inequality constraints are often reformulated into this form
%by the addition of slack variables, which we assume has already happened.
%
%General purpose open-source software is available for the solution of \eqref{eqn:nlopt}.
%In this paper, we use IPOPT \cite{ipopt}.
%As input, IPOPT requires oracles for
%the objective function $f(x): {\mathbb{R}}^N \rightarrow \mathbb{R}$, the gradient
%of the objective $\nabla f(x): {\mathbb{R}}^N \rightarrow \mathbb{R}^N$, the constraints
%$g(x): {\mathbb{R}}^N \rightarrow \mathbb{R}^M$, the Jacobian of the constraints
%$\nabla g(x): {\mathbb{R}}^N \rightarrow \mathbb{R}^{M\times N}$, and the \textit{Hessian-of-the-Lagrangian}
%$\nabla^2\mathcal{L}(x): {\mathbb{R}}^N \rightarrow \mathbb{R}^{N\times N}$, which is defined as:
%$$\nabla^2\mathcal{L}(x) = \sigma \nabla^2 f(x) + \sum\limits_{i=1}^M \lambda_i \nabla^2 g_i(x),$$
%where $\lambda$ is the Lagrangian multiplier associated with the equality constraint,
%and $\sigma$ is a scalar constant. Computing the first- and second-order derivatives
%of $f$ and $g$, and evaluating all the oracles is a potential bottleneck in the solution process.
%
We consider interior point methods, such as IPOPT \cite{ipopt}, for solving \eqref{eqn:nlopt}.
These methods iteratively compute search directions $d$ by solving the linear system
\eqref{eqn:linear_system}:
%With a slight abuse of notation to simplify unnecessary details, IPOPT's solution method
%for \eqref{eqn:nlopt} is an iterative algorithm that, at each iteration, involves the
%solution of the following linear system, which is related to the KKT conditions of \eqref{eqn:nlopt}:
\begin{equation}
  \begin{bmatrix}
    (\nabla^2\mathcal{L}(x)+\alpha) & \nabla g(x)^T \\
    \nabla g(x) & 0
  \end{bmatrix} d = -\begin{bmatrix}
      \nabla f(x) + \nabla g(x)^T \lambda + \beta\\
      g(x)
  \end{bmatrix},
  \label{eqn:linear_system}
\end{equation}
where $\alpha$ and $\beta$ are additional terms that are not shown for simplicity. The matrix on the left-hand side is referred to as the
{\it Karush-Kuhn Tucker}, or {\it KKT}, matrix.
%Solving this system, which
%is achieved through a factorization using third-party subroutines, is another potential
%bottleneck in the solution process.
To construct this system, solvers rely on callbacks that provide the Jacobian $\nabla g$
and, optionally, the Hessian of the Lagrangian function, $\nabla^2\mathcal{L}$. These are typically
provided by the automatic differentiation system of an algebraic modeling environment.
If the Hessian $\nabla^2\mathcal{L}$ is not available, a limited-memory quasi-Newton approximation
is used \cite{nocedal1980}.
%In this paper we use the Julia library JuMP \cite{jump1} and the nonlinear solver IPOPT
%\cite{ipopt} to build and solve all nonlinear optimization models. We use the
%MA27 linear solver from Harwell \cite{duff1983multifrontal} to solve the linear systems.

\subsection{Neural network predictors}
A neural network predictor is a function denoted $ y = {\rm NN}(x)$. We consider neural networks defined by repeated application of an affine transformation
and a nonlinear activation function $\sigma$ over $L$ layers:
\begin{equation}
  \begin{aligned}
    y_l &= \sigma_l (W_l y_{l-1} + b_l) && l \in \{1,\dots,L\},
    \label{eqn:nn}
  \end{aligned}
\end{equation}
where $y_0 = x$ and $y = y_L$.
Weights $W_l$ and biases $b_l$ are {\it parameters} that are optimized to minimize error
on a set of training data representing desired inputs and outputs of the neural network.
%and the constant data matrices $W_l$ and vectors $b_l$ are of the requisite dimension.
%The activation functions $\sigma_l$ are applied element-wise
%to the result of the affine transformation of each layer.
To a nonlinear optimization solver using a trained neural network, $W_l$ and $b_l$ are
constant.
To fit assumptions made by these solvers, we consider only smooth activation functions,
e.g., sigmoid and hyperbolic tangent functions.
%
%In this paper, we use the Python library PyTorch \cite{paszke2019pytorch} to model
%and train all neural networks.

\subsection{Algebraic representations of a neural network}\label{sec:algebraic}

In this section, we explain the three ways in which we encode pre-trained neural network
predictors into the constraints of a nonlinear optimization model of the form \eqref{eqn:nlopt}.
The three approaches are denominated \textit{full-space}, \textit{reduced-space}, and \textit{gray-box}.

\subsubsection{Full-space}

In the \textit{full-space} formulation, we add an intermediate vector-valued decision
variable $z_l$ to represent the output of the affine transformation in each layer
$l$, and we add a vector-valued decision variable $y_l$ to represent the output of
each nonlinear activation function. We then add a linear equality constraint
to enforce the relationship between $y_{l-1}$ and $z_l$ and a
nonlinear equality constraint to enforce the relationship between $z_l$ and $y_l$.
Thus, the neural network in \eqref{eqn:nn} is encoded by the constraints:
\begin{equation}
  \begin{aligned}
    {W_l} y_{l-1} - z_l &= -b_l && l \in \{1,\dots,L\} \\
    y_l - \sigma_l(z_l) &= 0 && l \in \{1,\dots,L\}.
    \label{eqn:nn_full}
  \end{aligned}
\end{equation}
The full-space approach prioritizes small expressions and small, sparse nonlinear constraints
at the cost of introducing additional variables and constraints for each layer of
the neural network.
This formulation conforms to the assumptions of JuMP's reverse-mode automatic differentiation
algorithm: 1) nonlinear constraints can be written as a set of scalar-valued functions,
and 2) they are sparse in the sense that each scalar constraint contains relatively
few variables and has a simple functional form.
%The benefits of the \textit{full-space} approach are that solvers and modeling layers
%can efficiently exploit the affine constraints, and that each nonlinear constraint
%is sparse (it uses only a small number of the total decision variables) and involves
%few nonlinear terms. The downside of the \textit{full-space} approach is that we add
%two scalar decision variables and two constraints for each output neuron of each layer.
%For deep (or wide) networks, this can result in a large number of additional variables
%and constraints in our nonlinear optimization model. Of particular relevance for our
%later computational results, this enlarges the size of the linear system we need to
%solve at each iteration.

\subsubsection{Reduced-space}
In the \textit{reduced-space} formulation we add a single
vector-valued decision variable $y$ to represent the output of the final activation
function, and we add a single vector-valued nonlinear equality constraint that encodes
the complete network.
Thus, the neural network in \eqref{eqn:nn} is encoded as the vector-valued constraint:
\begin{equation}
    y - \sigma_L({W_L} (\ldots\sigma_l({W_l} (\ldots\sigma_1({W}_1 x + b_1) \ldots) + b_l)\ldots) + b_L) = 0.
    \label{eqn:nn_reduced}
\end{equation}

The benefit of the \textit{reduced-space} approach is that we add only a single vector-valued
decision variable $y$ and a single vector-valued nonlinear equality constraint (each
of dimension of the output size of the last layer).
The downside is that the nonlinear constraint is a complicated expression with a very large
number of terms. This is made worse by the fact that JuMP scalarizes vector-valued
expressions in nonlinear constraints. Thus, instead of efficiently representing the
affine relationship ${W_1}x + b_1$ by storing the matrix and vector, JuMP instead represents
the expression as a sum of scalar products.

\subsubsection{Gray-box}

In the \textit{gray-box} formulation, we do not attempt to encode the neural network
algebraically. Instead we exploit the fact that nonlinear local solvers such as IPOPT
require only callback oracles to evaluate the constraint function $g(x)$ and the Jacobian
$\nabla g(x)$ (and, optionally, the Hessian $\nabla^2\mathcal{L}$). Using
JuMP's support for user-defined nonlinear operators, we implement the evaluation of
the neural network as a nonlinear operator ${\rm NN}(x)$, and we use PyTorch's built-in
automatic differentiation support to compute the Jacobian $\nabla {\rm NN}(x)$ and
Hessians $\nabla^2{\rm NN}(x)$. Thus, the neural network in \eqref{eqn:nn} is encoded
as the vector-valued constraint:
\begin{equation}
    y - {\rm NN}(x) = 0.
    \label{eqn:nn_gray}
\end{equation}

Like the reduced-space formulation, the gray-box approach adds only a small number
of variables and constraints to the optimization problem. By contrast, the gray-box
approach uses the automatic differentiation system of the neural network modeling
software, which is better-suited to the dense, nested, vector-valued expressions that
define the neural network. However, as explicit representation of the constraints
are not exposed to the solver, the gray-box formulation cannot support relaxations used
by nonlinear {\it global} optimization solvers.

\subsection{MathOptAI.jl}
Encoding a trained neural network into an optimization model using the forms described
in Section~\ref{sec:algebraic} is tedious and error-prone. To simplify experimentation,
we developed a new Julia package, MathOptAI.jl, which is a JuMP extension for embedding
a range of machine learning models into a JuMP model. In addition to supporting neural
networks trained using PyTorch, which are the focus of this paper, MathOptAI.jl also
supports Julia-based deep-learning libraries, as well as other machine learning models
such as decision trees and Gaussian Processes. MathOptAI.jl is provided as an open-source
package at \url{https://github.com/lanl-ansi/MathOptAI.jl} under a BSD-3 license.
%MathOptAI.jl is heavily inspired by existing software packages such as OMLT \cite{omlt}
% and gurobi-machinelearning \cite{gurobimachinelearning}.

\section{Test problem}
We compare the three different neural network formulations on a transient
stability-constrained, security-constrained ACOPF problem \cite{gan2000}.
\\

%\subsubsection{Stability-constrained optimal power flow}
{\bf Stability-constrained optimal power flow}~
Security-constrained optimal power flow (SCOPF) is a well-established problem
for dispatching generators in an electric power network in which feasibility
of the network (i.e., the ability to meet demand) is enforced for a set
of {\it contingencies} \cite{aravena2023}.
Each contingency $k$ represents the loss of a set of generators and/or power lines.
We consider a variant of this problem where, in addition to enforcing steady-state
feasibility, we enforce feasibility of the transient response resulting from
the contingency. In particular, we enforce that the transient frequency at each
bus is at least ${\bf \eta}=59.4$ Hz for the 30 second interval following each contingency.
This problem is given by Equation \ref{eqn:opf}:
\begin{equation}
  \min_{S^g,V} c(\mathbb{R}(S^g)) \text{ s.t. }
  \left\{\begin{array}{ll}
      F_k (S^g, V, {\bf S^d}) \leq 0 & k \in \{0,\dots,K\} \\
      G_k (S^g, {\bf S^d}) \geq {\bf \eta} \mathbbm{1} & k\in \{1,\dots,K\}.\\
  \end{array}\right.
  %\begin{array}{cll}
  %  \displaystyle \min_{S^g,V,S} & c(\mathbb{R}(S^g)) \\
  %  %\text{s.t.} & {\bf v^l} \leq \left| V_i \right| \leq {\bf v_i^u} & i\in N\\
  %  %& {\bf S^{gl}_i} \leq S_i^g \leq {\bf S_i^{gu}} & i \in N \\
  %  %& \left|S_{ij}\right| \leq {\bf S_{ij}^u} & (i,j) \in E\\
  %  %& \displaystyle S_i^g - {\bf S_i^d} = \sum_{(i,j)\in E\cup E^R} S_{ij} & \text i\in N \\
  %  %& S_{ij} = {\bf Y^*_{ij}} V_i V_i^* - {\bf Y^*_{ij}} V_i V_j^* & (i,j)\in E\cup E^R \\
  %  %& {\bf \theta^l_{ij}} \leq \angle (V_i V_j^*) \leq {\bf \theta^u_{ij}} & (i,j)\in E \\
  %  \text{s.t.} & F_k (S^g, V, {\bf S^d}) \leq 0 & k\in K \\
  %  & G_k (S^g, {\bf S^d}) \geq {\bf \eta} \mathbbm{1} & k\in K\\
  %\end{array}
  \label{eqn:opf}
\end{equation}
Here, $S^g$ is a vector of complex AC power generations for each generator in the network,
$V$ is a vector of complex bus voltages, $c$ is a quadratic cost function, and
$\bf S^d$ is a constant vector of complex power demands.
Here $F_k \leq 0$ represents the set of constraints enforcing feasibility of the
power network for contingency $k$, where $k=0$ refers to the base network,
and  $G_k$ maps generations and demands to the minimum frequency at each bus over the
interval considered.

In this work, we consider an instance of Problem \ref{eqn:opf} defined on a 37-bus
synthetic test network \cite{birchfield2017,hawaii40}.
In this case, $G_k$ has 117 inputs and 37 outputs.
We consider a single contingency that outages generator 5 on bus 23. We choose a small
network model with a single contingency because our goal is to test the different neural
network formulations, not the SCOPF formulation itself.
\\

%\subsubsection{Stability surrogate model}
{\bf Stability surrogate model}~
Instead of considering the differential equations describing transient behavior of the
power network directly in the optimization problem, we approximate $G_k$ with a neural
network trained on data from 110 high-fidelity simulations using PowerWorld \cite{powerworldmanual}
with generations and loads uniformly sampled from within a $\pm 20\%$ interval of each nominal value.
We use sequential neural networks with $\tanh$ activation functions with between
two and 20 layers and between 50 and 4,000 neurons per layer.
These networks have between 7,000 and 592 million trained parameters.
The networks are trained to minimize mean squared error using the Adam optimizer \cite{adam}
until training loss is below 0.01 for 1,000 consecutive epochs.
% TODO(rp): Table showing properties of each network (i.e. nodes, layers, parameters)
We use a simple training procedure and small amount of data because our goal is
to test optimization formulations with embedded neural networks, rather than the neural
networks themselves.

%Because the focus of this work is to test different formulations for trained neural
%networks in optimization problems, we do not focus extensively on validating the
%neural network model. In particular, we do not validate the behavior of the neural network
%surrogate on out-of-sample generation and load data.

\section{Results}

\subsection{Computational setting}
We model the SCOPF problem using PowerModels \cite{powermodels},
PowerModelsSecurityConstrained \cite{powermodelssecurity}, and JuMP \cite{jump1}.
Neural networks are modeled using PyTorch \cite{paszke2019pytorch} and embedded into
the optimization problem using MathOptAI.jl.
Optimization problems are solved using the IPOPT nonlinear optimization solver \cite{ipopt}
with MA27 \cite{duff1983multifrontal} as the linear solver.
The full-space and reduced-space models support evaluation on a CPU
but not on a GPU. Because gray-box models use PyTorch, they can be evaluated
on a CPU or GPU.
We run our experiments on the Venado supercomputer.
CPU-only experiments use compute nodes with two 3.4 GHz NVIDIA Grace CPUs and
240 GB of RAM, while
CPU+GPU experiments use nodes with a Grace CPU with 120 GB of RAM and an NVIDIA
H100 GPU. % with four sockets of 96 GB each.

\subsection{Structural results}

Table \ref{tab:structure} shows the numbers of variables, constraints, and
nonzero entries of the derivative matrices for the optimization problem with
different neural networks and formulations.
We note that the reduced-space and gray-box formulations have approximately the same
numbers of constraints and variables as the original problem, but more nonzero entries
in the Jacobian and Hessian matrices due to the dense, nonlinear stability constraints.
With these formulations, the structure of the optimization problem does not change
as the neural network surrogate adds more interior layers.
By contrast, the full-space formulation grows in numbers of variables, constraints,
and nonzeros as the neural network gets larger.
%We note that the number of nonzeros in the Jacobian scales with the number of trained
%parameters in the neural network.
These problem structures suggest that the full-space formulation will lead to expensive
KKT matrix factorizations, while this will not be an issue for reduced-space and gray-box
formulations.

\begin{table}%[h!]
  \centering
  \rowcolors{1}{gray!10}{white}
  \caption{Numbers of variables, constraints, and nonzeros for different networks and
  formulations}
  \begin{tabular}{cccccc}
    \hline\hline
    Parameters & Formulation   & N. Variables & N. Constraints & Jacobian NNZ & Hessian NNZ \\
    \hline
    --         & No surrogate  &         1155 &           1460 &         5822 &        1398 \\
    \hline
    7k         & Full-space    &         1292 &           1634 &        11796 &        1448 \\
    25k        & Full-space    &         1592 &           1934 &        27996 &        1598 \\
    578k       & Full-space    &         4192 &           4534 &       567896 &        2898 \\
    7M         & Full-space    &        17192 &          17534 &      7144896 &        9398 \\
    \hline
    All networks & Reduced-space &         1155 &           1497 &         8708 &        4479 \\
    %7k         & Reduced-space &         1155 &           1497 &         8708 &        4479 \\
    %25k        & Reduced-space &         1155 &           1497 &         8708 &        4479 \\
    %578        & Reduced-space &         1155 &           1497 &         8708 &        4479 \\
    %7M         & Reduced-space &         1155 &           1497 &         8708 &        4479 \\
    \hline
    All networks & Gray-box      &         1192 &           1534 &         8782 &        4479 \\
    %7k         & Gray-box      &         1192 &           1534 &         8782 &        4479 \\
    %25k        & Gray-box      &         1192 &           1534 &         8782 &        4479 \\
    %578        & Gray-box      &         1192 &           1534 &         8782 &        4479 \\
    %7M         & Gray-box      &         1192 &           1534 &         8782 &        4479 \\
    \hline\hline
  \end{tabular}
  \label{tab:structure}
\end{table}

\subsection{Runtime results}\label{sec:results}

\begin{table}[h!]
  \centering
  \rowcolors{1}{gray!10}{white}
  \caption{
  %{\color{red} Robby: I played around changing the order of the rows.}
  Solve times with different neural networks and formulations}
  \resizebox{\textwidth}{!}{
  \begin{tabular}{cccccccc}
    \hline\hline
    Parameters & Formulation   & Hessian         & Platform & Build time & Solve time & Iterations & Time/iter. \\
    \hline
    --         & No surrogate  & Exact           & CPU      &      45 ms &      0.4 s &         41 &       9 ms \\
    \hline
    7k         & Full-space    & Exact           & CPU      &      0.1 s &        2 s &        468 &       4 ms \\
    25k        & Full-space    & Exact           & CPU      &      0.3 s &        5 s &        642 &       8 ms \\
    578k       & Full-space    & Exact           & CPU      &      0.3 s &      699 s &        755 &      0.9 s \\
    7M         & Full-space$^*$& Exact           & CPU      &      --    &      --    &        --  &       --   \\
    \hline
    7k         & Reduced-space & Exact           & CPU      &      0.1 s &        7 s &         49 &      0.1 s \\
    25k        & Reduced-space & Exact           & CPU      &        1 s &     1125 s &         41 &       27 s \\
    578k       & Reduced-space$^\dag$  & Exact   & CPU      &      --    &     --     &         -- &      --    \\
    7M         & Reduced-space$^\ddag$ & Exact   & CPU      &        --  &     --     &         -- &      --    \\
    \hline
    7k         & Gray-box      & Exact           & CPU      &      0.1 s &        8 s &         41 &      0.2 s \\
    25k        & Gray-box      & Exact           & CPU      &      0.1 s &        9 s &         42 &      0.2 s \\
    578k       & Gray-box      & Exact           & CPU      &      0.1 s &       11 s &         42 &      0.3 s \\
    7M         & Gray-box      & Exact           & CPU      &      0.1 s &       22 s &         42 &      0.5 s \\
    68M        & Gray-box      & Exact           & CPU      &      0.1 s &      140 s &         42 &        3 s \\
    592M       & Gray-box      & Exact           & CPU      &      0.6 s &      748 s &         42 &       18 s \\
    \hline
    7k         & Gray-box      & Exact           & CPU+GPU  &      0.1 s &        7 s &         41 &      0.2 s \\
    25k        & Gray-box      & Exact           & CPU+GPU  &      0.1 s &        7 s &         42 &      0.2 s \\
    578k       & Gray-box      & Exact           & CPU+GPU  &      0.1 s &        8 s &         42 &      0.2 s \\
    7M         & Gray-box      & Exact           & CPU+GPU  &      0.1 s &        8 s &         42 &      0.2 s \\
    68M        & Gray-box      & Exact           & CPU+GPU  &      0.2 s &        9 s &         42 &      0.2 s \\
    592M       & Gray-box      & Exact           & CPU+GPU  &      0.7 s &       15 s &         42 &      0.3 s \\
    \hline
    %7k         & Full-space    & Approx.         & CPU      &      0.1 s &      0.7 s &        199 &       4 ms \\
    %25k        & Full-space    & Approx.         & CPU      &      0.1 s &        5 s &        565 &       8 ms \\
    %578k       & Full-space    & Approx.         & CPU      &      0.5 s &      607 s &        734 &      0.8 s \\
    %7M         & Full-space$^*$& Approx.         & CPU      &      --    &      --    &        --  &       --   \\
    %\hline
    %7k         & Reduced-space & Approx.         & CPU      &      0.1 s &      0.7 s &         74 &       7 ms \\
    %25k        & Reduced-space & Approx.         & CPU      &        2 s &       50 s &         58 &      0.8 s \\
    %578k       & Reduced-space$^\dag$  & Approx. & CPU      &      --    &     --     &         -- &      --    \\
    %7M         & Reduced-space$^\ddag$ & Approx. & CPU      &        --  &     --     &         -- &      --    \\
    %\hline
    7k         & Gray-box      & Approx.         & CPU      &      0.1 s &      0.3 s &         61 &       6 ms \\
    25k        & Gray-box      & Approx.         & CPU      &      48 ms &      0.3 s &         57 &       6 ms \\
    578k       & Gray-box      & Approx.         & CPU      &      0.1 s &        1 s &         66 &      15 ms \\
    7M         & Gray-box      & Approx.         & CPU      &      0.1 s &        6 s &         57 &      0.1 s \\
    68M        & Gray-box      & Approx.         & CPU      &      0.1 s &        7 s &         56 &      0.1 s \\
    592M       & Gray-box      & Approx.         & CPU      &      0.9 s &       17 s &         56 &      0.3 s \\
    \hline
    7k         & Gray-box      & Approx.         & CPU+GPU  &      50 ms &      0.5 s &         63 &       7 ms \\
    25k        & Gray-box      & Approx.         & CPU+GPU  &      48 ms &      0.4 s &         58 &       7 ms \\
    578k       & Gray-box      & Approx.         & CPU+GPU  &      0.1 s &      0.5 s &         62 &       8 ms \\
    7M         & Gray-box      & Approx.         & CPU+GPU  &      0.1 s &      0.5 s &         57 &       9 ms \\
    68M        & Gray-box      & Approx.         & CPU+GPU  &      0.2 s &        1 s &         56 &      21 ms \\
    592M       & Gray-box      & Approx.         & CPU+GPU  &      0.7 s &        1 s &         56 &      23 ms \\
    \hline\hline
    \rowcolor{white}\multicolumn{8}{l}{\scriptsize $^*$ Fails with a segfault, possibly due to memory requirements of MA27}\\[-2pt]
    \rowcolor{white}\multicolumn{8}{l}{\scriptsize $^\dag$ Exceeds resource manager's memory limits}\\[-2pt]
    \rowcolor{white}\multicolumn{8}{l}{\scriptsize $^\ddag$ Exceeds ten-hour time limit}\\
  \end{tabular}
  }
  \label{tab:alltimes}
\end{table}

Runtimes for the different formulations with neural network surrogates of increasing size
are given in Table \ref{tab:alltimes}.
For gray-box formulations, we compare optimization solves using exact and approximate
Hessian evaluations and different hardware platforms.
The results immediately show that full-space and reduced-space formulations are not scalable to
neural networks with more than a few million trained parameters. The full-space formulation
fails with a segmentation fault---likely due to memory issues in MA27---while the reduced-space
formulation exceeds time and memory limits building the constraint expressions in JuMP.
A breakdown of solve times, given in Table \ref{tab:solvetime-breakdown}, confirms the
bottlenecks in these formulations. The full-space formulation spends almost all of its
solve time in the IPOPT algorithm, which we assume is dominated by KKT matrix factorization,
while the reduced-space formulation spends most of its solve time evaluating the Hessian.
%When an exact Hessian is used, the reduced-space formulation spends almost all of
%its solve time evaluating this Hessian. When an approximate Hessian is used, the solve time is
%dominated by function evaluation.

By contrast, the gray-box formulation is capable of solving the optimization problem
with the largest neural network surrogates tested.
While a CPU-only solve with exact Hessian matrices takes an unacceptably-long 748 s,
a GPU-accelerated solve with approximate Hessian matrices solves in only one second.
This is slower than the original SCOPF problem (with no stability constraint) by a
factor of 2.5, which may be acceptable for some applications.
In all cases, the solve time with the gray-box formulation is dominated by function
and Hessian evaluations, which explains the large speed-ups obtained with the GPU
(50$\times$ and 17$\times$ for the 592M-parameter network with exact and approximate Hessians).
%In fact, the approximate-Hessian solve time with GPU acceleration is still dominated by function
%evaluation, suggesting that further speed-ups would be possible with more powerful hardware.

Approximating the Hessian matrix also speeds up the solve significantly.
Hessian approximation is not a common approach when exact Hessians are available
because it can lead to slow and unreliable convergence. In the 592M-parameter case,
approximating the Hessian increases the iteration count by 14, but it makes up for this
by decreasing the time per iteration by a factor of 60.
These results suggest that this is an appropriate trade-off for optimization problems
constrained by large neural networks.

\begin{table}[h!]
  \centering
  \caption{Solve time breakdowns for selected neural networks and formulations}
  \rowcolors{3}{gray!10}{white}
  \resizebox{\textwidth}{!}{
  \begin{tabular}{ccccc|ccccc}
    \hline\hline
    \multirow{2}{*}{Formulation} &\multirow{2}{*}{Parameters} &\multirow{2}{*}{Hessian} &\multirow{2}{*}{Platform} &\multirow{2}{*}{Solve time}& \multicolumn{5}{c}{Percent of solve time (\%)} \\
    & & & & & Function & Jacobian & Hessian & Solver & Other \\
    \hline
    Full-space    & 578k       & Exact           & CPU      &      699 s &       0.1 &    $<$0.1 &       0.2 &       99+ &    $<$0.1 \\
    %Full-space    & 578k       & Approx.         & CPU      &      607 s &       0.1 &    $<$0.1 &        -- &       99+ &    $<$0.1 \\
    %\hline
    Reduced-space & 25k        & Exact           & CPU      &     1125 s &         2 &       0.5 &        97 &       0.4 &       0.3 \\
    %Reduced-space & 25k        & Approx.         & CPU      &       50 s &        68 &        15 &        -- &        10 &         8 \\
    %\hline
    Gray-box      & 592M       & Exact           & CPU      &      748 s &        97 &         2 &         1 &    $<$0.1 &    $<$0.1 \\
    Gray-box      & 592M       & Exact           & CPU+GPU  &       15 s &        54 &         1 &        42 &         2 &       0.6 \\
    Gray-box      & 592M       & Approx.         & CPU      &       17 s &        96 &         2 &        -- &         2 &    $<$0.1 \\
    Gray-box      & 592M       & Approx.         & CPU+GPU  &        1 s &        76 &         6 &        -- &        17 &       0.1 \\
    \hline\hline
  \end{tabular}
  }
  \label{tab:solvetime-breakdown}
\end{table}

\section{Conclusion}

This work demonstrates that nonlinear local optimization problems may incorporate neural
networks with hundreds of millions of trained parameters, with modest overhead,
using a gray-box formulation that exploits efficient automatic differentiation, Hessian approximation, and
and GPU acceleration. A disadvantage of the gray-box formulation is that it is not suitable
for global optimization as the non-convex neural network constraints cannot be relaxed.
Additionally, relative performance of the formulations may change in different applications.
This motivates future research and development to improve the performance
of the full-space and reduced-space formulations.
The full-space formulation may be improved by decomposing the KKT matrix
to exploit the structure of the neural network's Jacobian,
while the reduced-space formulation may be improved by exploiting vector-valued functions
and common subexpressions in JuMP.

%\begin{credits}
%\subsubsection{\ackname} A bold run-in heading in small font size at the end of the paper is
%used for general acknowledgments, for example: This study was funded
%by X (grant number Y).
%
%\subsubsection{\discintname}
%It is now necessary to declare any competing interests or to specifically
%state that the authors have no competing interests. Please place the
%statement with a bold run-in heading in small font size beneath the
%(optional) acknowledgments\footnote{If EquinOCS, our proceedings submission
%system, is used, then the disclaimer can be provided directly in the system.},
%for example: The authors have no competing interests to declare that are
%relevant to the content of this article. Or: Author A has received research
%grants from Company W. Author B has received a speaker honorarium from
%Company X and owns stock in Company Y. Author C is a member of committee Z.
%\end{credits}
\bibliographystyle{splncs04}
\bibliography{ref}
\end{document}